\documentclass[11pt, oneside]{article}

\usepackage[final,nonatbib]{neurips_2022}

\usepackage{float}
\usepackage{geometry}
\usepackage{subcaption}
\usepackage{graphicx}
\usepackage{color}
\usepackage{soul}
\usepackage{url}
\usepackage{amsmath}
\usepackage{xcolor}
\usepackage{amssymb}
\usepackage{hyperref}
\usepackage{authblk}

\title{NeurIPS 2022 Competition:\\Driving SMARTS}

\author[]{Amir Rasouli}
\author[]{Randy Goebel}
\author[]{Matthew E. Taylor}
\author[]{Iuliia Kotseruba}
\author[]{Soheil Alizadeh}
\author[]{Tianpei Yang}
\author[]{Montgomery Alban}
\author[]{Florian Shkurti}
\author[]{Yuzheng Zhuang}
\author[]{Adam Scibior}
\author[]{Kasra Rezaee}
\author[]{Animesh Garg}
\author[]{David Meger}
\author[]{Jun Luo}
\author[]{Liam Paull}
\author[]{Weinan Zhang}
\author[]{Xinyu Wang}
\author[]{Xi Chen}

\affil[]{{\texttt{smarts4ad@gmail.com}}}

\begin{document}
\maketitle

\begin{abstract}
Driving SMARTS is a regular competition designed to tackle problems caused by the distribution shift in dynamic interaction contexts that are prevalent in real-world autonomous driving (AD). The proposed competition supports methodologically diverse solutions, such as reinforcement learning (RL) and offline learning methods, trained on a combination of naturalistic AD data and open-source simulation platform SMARTS. 

The two-track structure allows focusing on different aspects of the distribution shift. Track 1 is open to any method and will give ML researchers with different backgrounds an opportunity to solve a real-world autonomous driving challenge. Track 2 is designed for strictly offline learning methods. Therefore, direct comparisons can be made between different methods with the aim to identify new promising research directions. 

The proposed setup consists of 1) realistic traffic generated using real-world data and micro simulators to ensure fidelity of the scenarios, 2) framework accommodating diverse methods for solving the problem, and 3) baseline method. As such it provides a unique opportunity for the principled investigation into various aspects of autonomous vehicle deployment.

\end{abstract}

\subsection*{Keywords}
Autonomous driving, distribution shift, real-world data, reinforcement learning.
\subsection*{Competition type}
Regular

\section{Competition description}

\subsection{Background and impact}

Through the proposed Driving SMARTS competition, we aim to attract attention of the machine learning (ML) community to challenges resulting from distribution shift in diverse dynamic interaction contexts that are prevalent in real-world autonomous driving (AD). Regardless of the solution methods, be it reinforcement learning (RL), supervised learning, or classical planning and control, such challenges must be adequately addressed to allow broader deployment of autonomous vehicles and to continue principled investigation into the opportunities and dangers of such deployment in an open social environment of our streets and roads. 

The Driving SMARTS competition is designed to have a broad appeal to diverse expertise in the NeurIPS community by supporting a wide range of solutions, such as single- and multi-agent RL, supervised learning, and control and optimization methods with data-driven enhancements. Competition scenarios and evaluation schemes are set up to allow all such methods to show their unique strengths. Moreover, the competition stays close to real-world challenges by emphasizing use of naturalistic driving data, fidelity of scenario design, and relevance of evaluation metrics.
 
To raise the scientific question of distribution shift in autonomous driving interaction to the broadest audience and emphasize an important recent technical direction, we adopt a two-track structure. The first, \textbf{freestyle} track is open to any method, allowing each to show its level of performance. Working on this track will give the participants a direct sense of the real-world autonomous driving challenges regardless of their expertise domain. Together, the participants' contributions will give the ML4AD community a clear understanding of how these various approaches fare against the challenge of distribution shift under interaction. The second, \textbf{offline} track, is designed for offline learning methods and helps answer important questions such as: How do offline methods stack up against other approaches? How far can we push offline learning? How could offline-to-online distribution shift be adequately dealt with for real-world deployment? How could distribution shift induced by multiple AI agents (e.g. when multiple autonomous cars meet) be dealt with? By using the same set of tasks across two tracks we will be able to directly compare offline methods with semi-offline and non-offline ones. This will help the community understand the shape of the "envelope" of offline methods and identify directions for further research.

Driving SMARTS puts heavy emphasis on bringing challenges of real-world autonomous driving R\&D to the doorstep of the ML researchers such that they can devise innovative solutions without being entangled in the complex practicalities of data preprocessing, scenario construction, evaluation, and real-world system integration. To this end, we propose meaningful scenarios integrated within a mature simulation platform. These scenarios include multi-lane cruising with specific target lane selection and the need for lane change, on-ramp merge where an autonomous vehicle has to join a flow of background traffic with subsequent lane changes in short sequence, and turns without traffic light protection. Where possible, we include naturalistic driving data for such scenarios so that offline methods could be used to enhance the solution's ability, realism, and real-world relevance. All data and scenarios will be available through the award-winning (CoRL 2020 Best System Paper) open source simulation and ML4AD platform SMARTS\footnote{\url{https://github.com/huawei-noah/SMARTS/}} \cite{zhou2020smarts}. Real-world AD data can be downloaded from the original source and used in SMARTS through preprocessing scripts provided by the organizers. The competition will be hosted on a free and open competition hosting platform CodaLab\footnote{\url{https://codalab.org/}}.

\subsection{Novelty}

Autonomous driving challenges held over the past few years focused predominantly on perception (detection and tracking of objects/events) and prediction (single- and multi-agent future trajectory prediction given observations). In our proposal we instead aim at planning and control. Among the competitions that targeted similar problems are CARLA Autonomous Driving Challenge\footnote{\url{https://carlachallenge.org/}} held in 2019 and SMARTS Autonomous Driving Competition organized by our colleagues during the Distributed AI (DAI) conference in 2020. CARLA AD Challenge consisted of predefined routes with multiple scenarios based on the NHTSA pre-crash typology with tracks for different perception stacks and evaluation focused on traffic infractions. SMARTS AD competition focused on planning and interaction in a single- and multi-agent settings across a variety of everyday driving scenarios (on-ramp, off-ramp, double merge, T-junction, etc.).

The proposed Driving SMARTS competition offers the following novel features:
\begin{itemize}
\itemsep0em
\item Use of naturalistic driving data in combination with simulation platform to produce novel scenarios with interactive traffic;
\item Offline learning track for direct comparisons between different learning methods on the same tasks.
\end{itemize}

\subsection{Data}

Training and evaluation data for the competition will consist of diverse scenes from open-source naturalistic driving datasets (e.g. NGSIM \cite{kovvali2007video},  Waymo \cite{Sun_2020_CVPR},) on roads and highways as well as  synthetic data. The scenarios vary in terms of number of lanes, traffic density, and speed limit. The driving tasks include slowing down, overtaking, changing lanes, turning at the intersections, etc. 

These scenarios will be replayed through SMARTS simulation platform that supports construction of environments and control of agents to emulate real-world driving experience at different levels of granularity. SMARTS allows exploring cases that are not included in the real-world driving datasets. To this end, the scenarios mined from large-scale naturalistic public datasets will be manipulated and combined to create unseen test scenes for the competition. Finally, in some scenarios, SMARTS will control interactive background vehicles.

\subsection{Tasks and application scenarios}
\subsubsection{Competition tracks}
\label{sec:competition_tracks}

There are two tracks in the competition with identical tasks and evaluation metrics but different training methods as described below.
\begin{description}
\itemsep0em
\item \textbf{Track 1:} The participants may use any method and training data to develop their solutions to complete the tasks. 

\item \textbf{Track 2:} This track targets the offline learning capacity of the methods. Only methods trained solely on offline datasets are accepted.

\end{description}
Participants will be invited to submit their complete training and inference code for evaluation. Here, only offline training methods are accepted and will be eligible for Track 2. The methods will be trained and tested by the organizers on an unseen set of train and test data. The participants may opt out of participating in  Track 2.
\subsubsection{Tasks}

\label{sec:tasks}
In both tracks, the participants are expected to successfully complete the given tasks. The goal for the agent is to drive as quickly and safely as possible from the start line to a pre-defined finish location, amid background traffic that consists only of vehicles. 

Participants are expected to develop a parameter-sharing multi-agent model that can control a single agent or a group of agents to accomplish specific tasks described below.

\begin{figure}[H]
\centering
\subfloat[][Multi-agent cruising]{\includegraphics[width=0.6\textwidth]{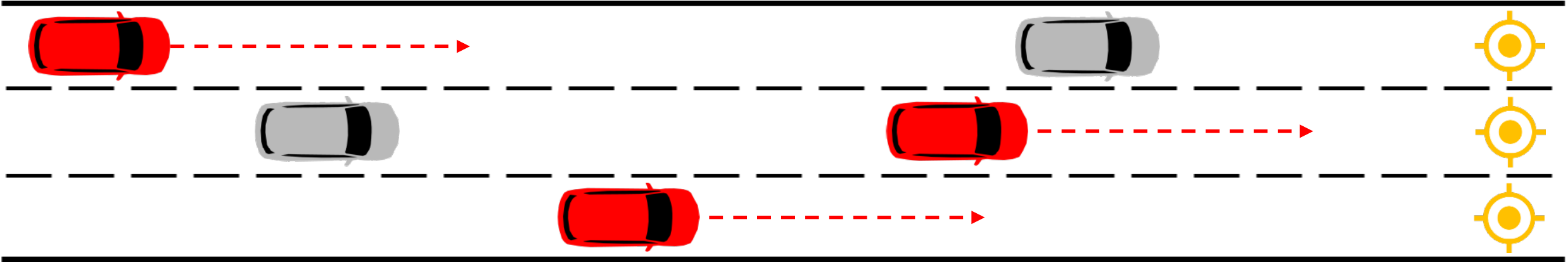}}\\
\subfloat[][Multi-agent merging]{\includegraphics[width=0.6\textwidth]{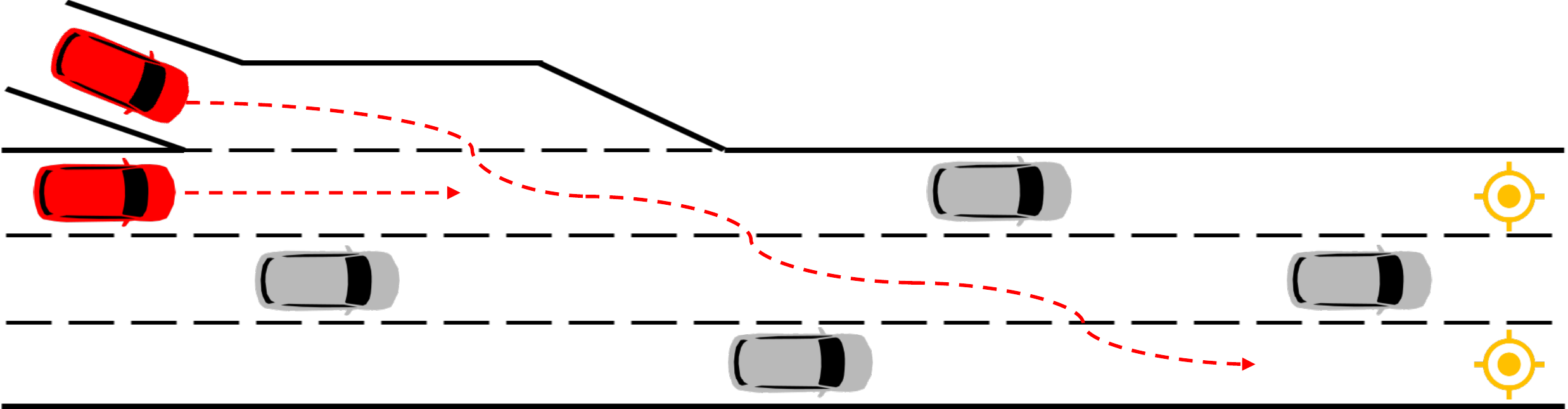}}\\
\begin{subfigure}[b]{0.35\textwidth}
\includegraphics[width=\textwidth]{images/caption.pdf}
\end{subfigure}
\caption{Multi-agent tasks}
\label{fig:multi_agent_tasks}
\end{figure}

\begin{itemize}
\itemsep0em
\item \textbf{Single-agent tasks} (Figure \ref{fig:single_agent_static_target}):
\begin{itemize}

\item \textit{Cruising} - an entry-level task to check the stability and human-like capabilities of the method. In this task, the vehicle has to safely cruise to a specific location without the need for any specific maneuvers. 
\item \textit{Overtaking} - tests the capability of the agent to maneuver among slow traffic to decrease its travel time.
\item \textit{Merging} - the agent has limited time to perform a lane changing maneuver to merge into the traffic in another lane, otherwise an accident may occur. 
\item\textit{Cutting off} - the agent is cut off aggressively by another vehicle and has to a take proper action to avoid an accident (e.g. by braking or changing lane).
\item \textit{Left turn at unsignalized T-intersection} - tests the solution on various criteria, such as following the stop sign, estimating time to cross the intersection, turning into the correct lane and accelerating without causing accidents.
\item \textit{left/right turn at unsignalized intersection }- this task requires the agent to be aware of the traffic moving in the opposite direction and encourages riskier maneuvers that might cause reactions from background vehicles.
\end{itemize}

\begin{figure}[H]
\centering
\subfloat[][Cruising]{\includegraphics[width=0.6\textwidth]{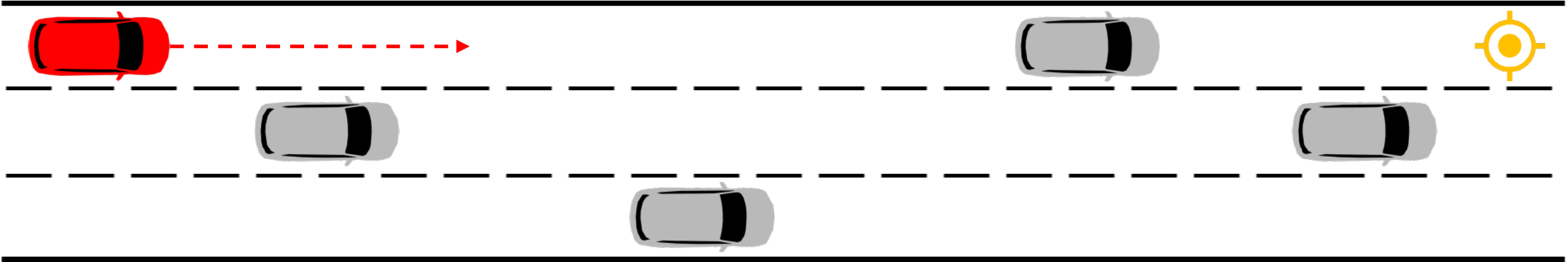}}\\
\subfloat[][Overtaking]{\includegraphics[width=0.6\textwidth]{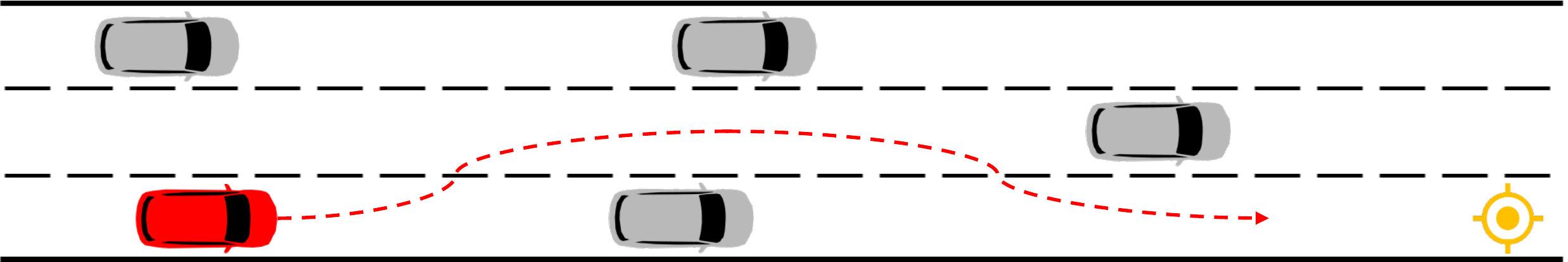}}\\
\subfloat[][Merging]{\includegraphics[width=0.6\textwidth]{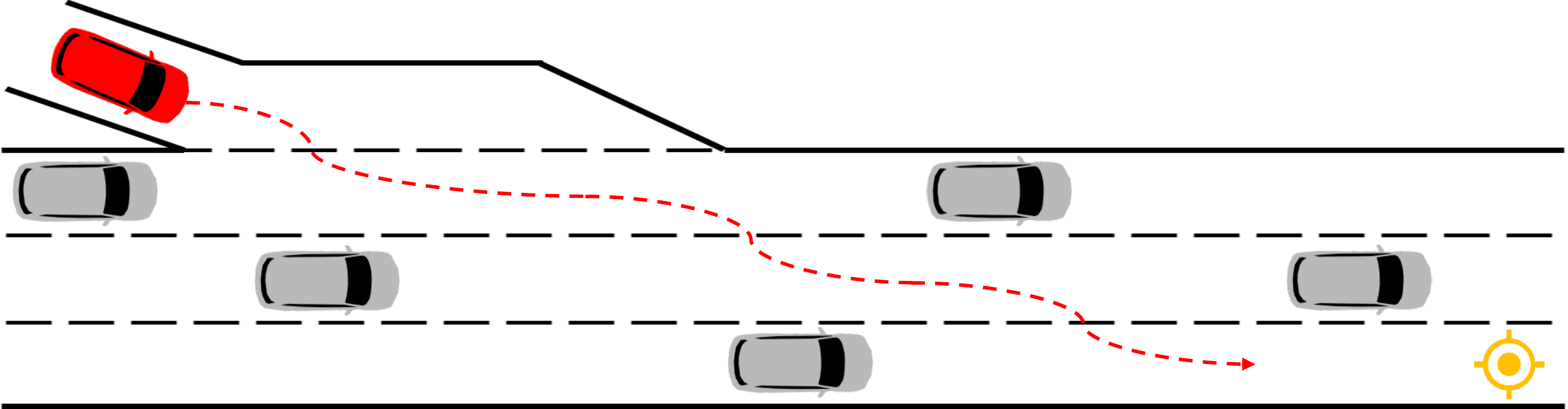}}\\
\subfloat[][Cutting off]{\includegraphics[width=0.6\textwidth]{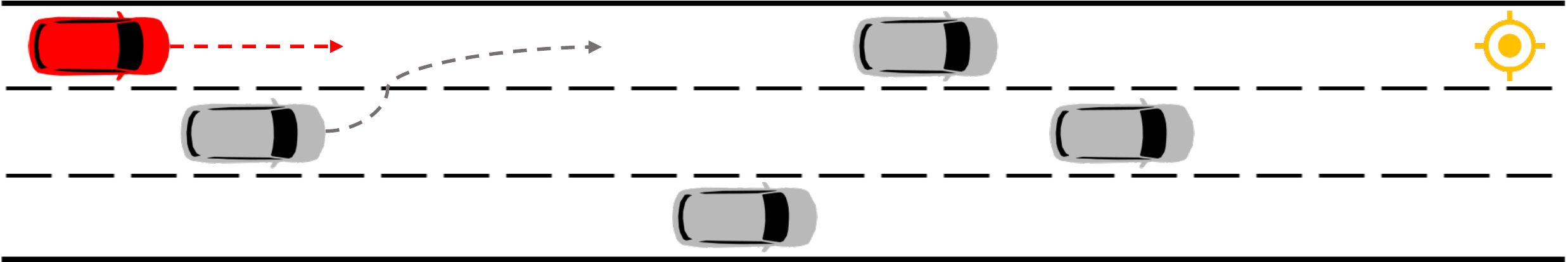}}\\
\subfloat[][Turn at T-intersection]{\includegraphics[width=0.6\textwidth]{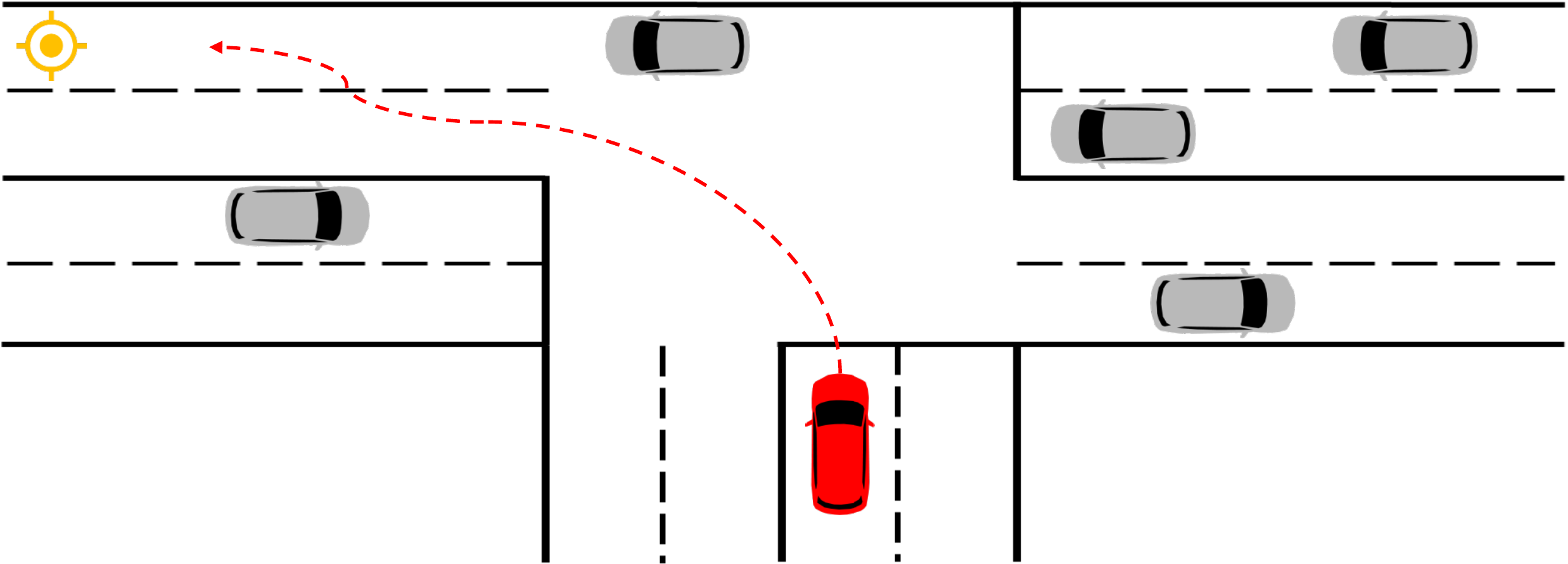}}\\
\subfloat[][Turn at intersection]{\includegraphics[width=0.6\textwidth]{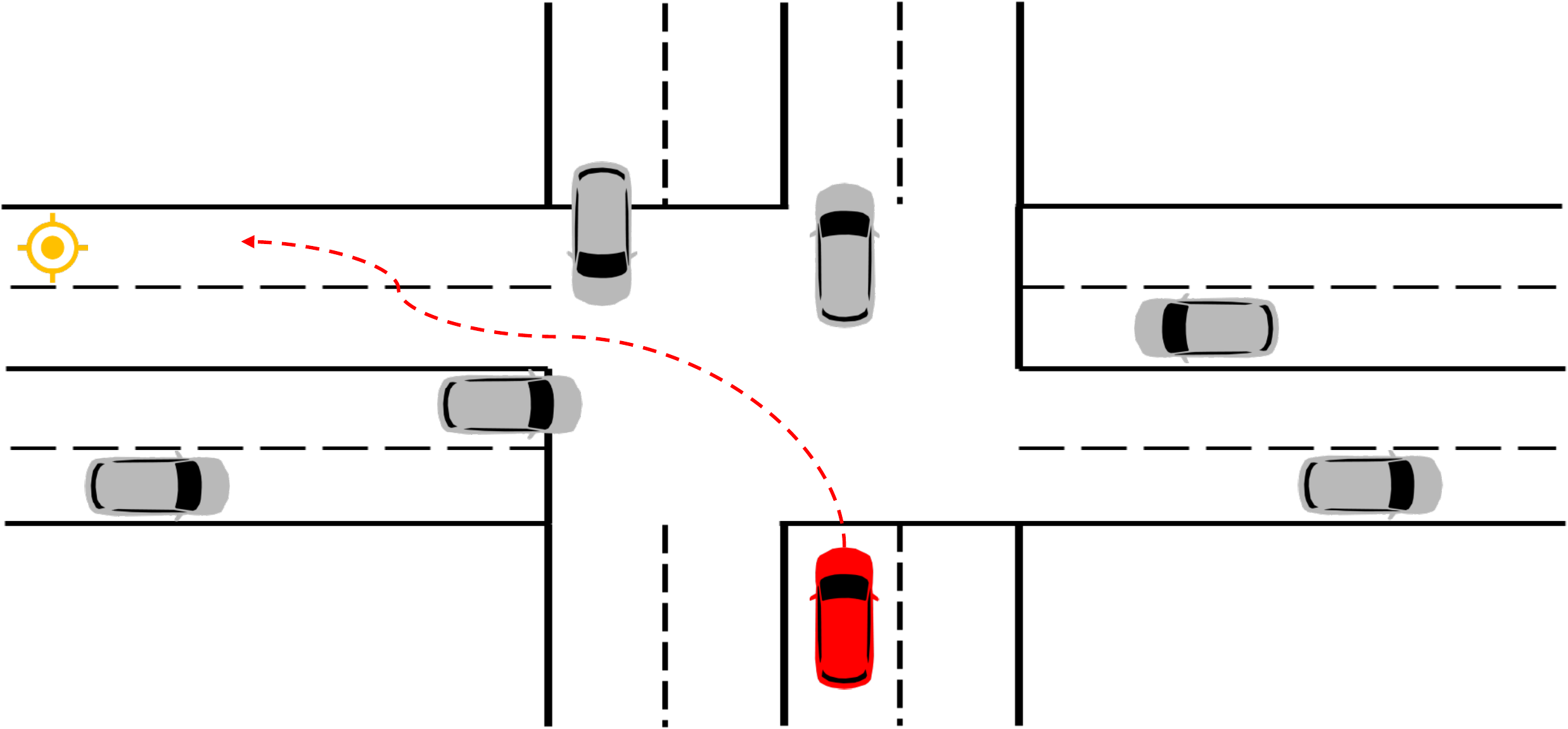}}\\

\begin{subfigure}[b]{0.35\textwidth}
\centering
\includegraphics[width=\textwidth]{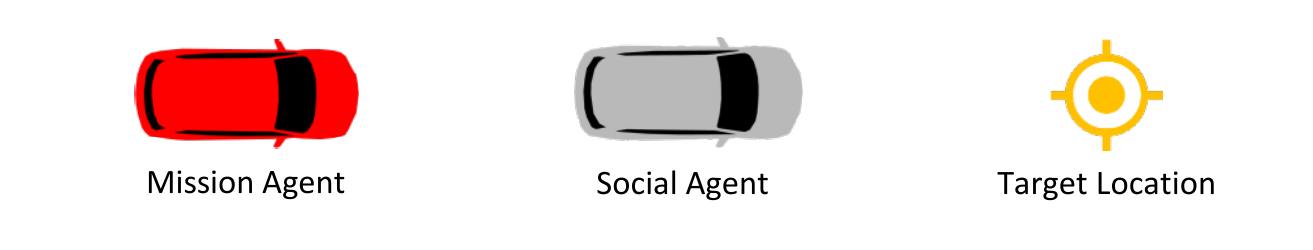}
\end{subfigure}
\caption{Single-agent tasks}
\label{fig:single_agent_static_target}
\end{figure}

\item \textbf{Multi-agent tasks} (Figure \ref{fig:multi_agent_tasks}):
\begin{itemize}
\item \textit{Cruising} - this task is similar to single-agent case but controlling multiple agents at the same time.
\item \textit{Merging} - requires two agents to coordinate to complete the merging task safely and quickly. Two agents must interact closely under time constraints.
\end{itemize}
\end{itemize}

The background traffic in all tasks are either reactive or passive (i.e. does not react to the controlled agent). Lack of interference from other agents increases the stability and safety of the solutions and in some cases can resemble behaviors of distracted drivers.

Each task will have a number of scenarios with different road and traffic configurations.

\subsection{Metrics}

Four metrics are used to evaluate the methods. The metrics listed below in the order of importance are used to rank the methods. In case of a tie, the second metric is used as the tie breaker. For all metrics, smaller value is better. 

\begin{enumerate}
\itemsep0em

\item \textbf{Completion}: This metric captures how many scenarios have been successfully completed, or in this case in how many cases the agent failed to accomplish its goal. To convert it into an error measure, the metric is computed as follows: 

$$\mathtt{completion}=\frac{\mathtt{num.\ failures}}{\mathtt{num.\ sc}}$$
\noindent where $sc$ is $\mathtt{scenario}$.

\item \textbf{Time}: The average time it takes for the agent to finish a scenario. This metric meant to capture the efficiency of a path generated by an algorithm:
$$ \mathtt{time} = \frac{\sum_{sc}T^i}{\mathtt{num. \ sc}}$$
\noindent where $T^i$ is the number of time steps for scenario $i$. 

\item \textbf{Humanness}: The agents should display driving skills similar to human drivers. For example, frequent direction changes, hard acceleration and braking might score higher on some metrics but are not human-like. This metrics computed as follows:

$$\mathtt{humanness} = \frac{\sum_T \mathtt{distance}_{obs} + \mathtt{jerk} +
\mathtt{steer}_{rate} + lc_{off} }
{\sum_{sc}T^i}$$ 
\noindent $obs$ is obstacle, $lc_{off}$ is lane center offset given by,
$$lc_{off} = (\frac{\mathtt{distance \ from \ center}}{\mathtt{lane \ width}})^2.$$

\item \textbf{Rule violations}: The agents should drive safely by following the traffic laws. We consider two violations, namely exceeding speed limits and driving in the wrong direction:

$$\mathtt{rule}= \frac{\sum_{sc}\sum_{T^i}
min(\frac{s_{\mathtt{violate}}}{0.5\times s_{\mathtt{limit}}}, 1) +
\mathtt{road}_{\mathtt{violate}}}{\sum_{sc}T^i}$$

\noindent where $s_{\mathtt{violate}} = max(0, s_a - s_{limit})$ is amount of violating speed (i.e. speed over the limit) and $\mathtt{road}_{\mathtt{violate}} \in \{0,1\}$ is whether road direction violated. 

\end{enumerate}

\subsection{Baseline, code, and materials provided}
\label{sec:materials}

\textbf{Baseline and code.} Participants will be provided with a starter kit that can be downloaded from the competition site. The kit will include baseline example code, evaluation code, details of evaluation metrics, a small validation set, and instructions required to make a submission. An image of a working participant environment will also be made available to alleviate software issues. 

The baseline will demonstrate interface usage and give examples of approaches that participants can use for their solutions. We will provide a simple RL-based baseline with code for inference and evaluation.

\textbf{Hardware.} This competition will be hosted remotely and contains no hardware platform materials.

\textbf{Competition site and leaderboard.} Submissions will be made to the competition site with the public leaderboard hosted on CodaLab. The leaderboard will show the ranks of entries as well as the contributions of individual metrics to the overall score. Evaluation results from the baseline will be entered into the leaderboard at the start of the competition as a reference for participants.

\subsection{Website, tutorial and documentation}

Documentation include instructions on how to acquire and adapt external datasets and details of the interface for the submission site and code. Specific details of the data format and samples for various tasks are provided. External documentation is referenced when necessary. All documentation is available at \url{https://smarts-project.github.io}.

\section{Organizational aspects}
\subsection{Protocol}
\label{sec:protocol}

\textbf{How to enter:}
The competition starts on August 1, 2022 and ends on November 7, 2022. Any submission after the end date will be automatically disqualified. The following describes how entries to the competition can be submitted by the individual participant or team representative (the participant):

\begin{itemize}
\itemsep0em
\item The participant (single entrant or representative of the team) must register on the official competition website before the end date of the competition;
\item The participant will be provided with login details to the third-party evaluation system CodaLab to gain access to the competition and submit a competition entry. The use of the third party evaluation system is subject to the Privacy Policy and Terms of Use of the operator of that system;
\item The required documentation and a starter kit (as described in Section \ref{sec:materials}) will be hosted on the competition website and accessible upon registration;
\item The entry submitted by the participant will consist of the pretrained model and inference code. Select participants who opted to compete in Track 2 will be requested to submit their training code.
\end{itemize}

\textbf{Validation:}
\begin{itemize}
\itemsep0em

\item Submissions for Track 1 will be evaluated automatically on the withheld test set and evaluation results will be available on the public leaderboard hosted on Codalab. 

\item Participants in Track 1 who used offline methods are invited to submit their code for consideration in track 2. At this stage up to 6 methods are used based on their ranking in Track 1. The organizers will train and evaluate the models and inspect the submitted code.

\item The organizers will work with participants who selected Track 2 to run their training code to retrain agents. Participants may be asked to submit additional documentation at this stage.

\item The winners of Track 1 will be determined based on the overall performance on the test set.

\item The winners of Track 2 will be determined based on the results obtained using the models trained by the organizers.

\item Since winners of Track 2 are a subset of Track 1 participants, the same entry may be eligible for cash prizes in both tracks.

\end{itemize}

\textbf{Cheating and overfitting prevention:}
\begin{itemize}
\itemsep0em
\item The participants in Track 1 will not have access to the test set used for evaluation.
\item The participants in Track 2 will not have access to the training and test sets used for evaluation.
\item The participants will be allowed up to 6 submission attempts.
\item The organizers will inspect the inference and training code of the finalists.
\end{itemize}

\subsection{Rules}

To be eligible for receiving cash prizes the participants must satisfy the following criteria:

\begin{itemize}
\itemsep0em
\item Participants must be at least 18 years of age.
\item Not be an organizer of the competition nor a family member of a competition organizer.
\item Participants can be a single entrant or part of the team.
\item Participants must be solely responsible for creating the submission. Receiving supervision or advice from the participants' supervisors or colleagues is permitted.
\item Participants must be registered for the competition with their email address before submitting an entry.
\item Entries must be open-source, participants are responsible for the license of the data they use for developing their models.
\end{itemize}

\subsection{Schedule and readiness}
The following is the schedule for the conference duration:
\begin{itemize}
\itemsep0em
\item Aug. 1, 2022: Competition opens
\item Nov. 7, 2022: Competition closes
\item Nov. 12, 2022: Deadline for participants to submit their code for Track 2
\item Nov. 20, 2022: Winning teams will be announced
\item Nov. 20, 2022: Winning teams are invited to submit a short report of their proposed approach.
\end{itemize}

The participants will be provided with all required materials needed for entering the competition including the software and detailed instructions of the competition. There is no requirement for pre-registration into the competition and there is no limit on how many teams can compete.

\section{Resources}
\subsection{Organizing team}

The organizing team is comprised of a diverse group of professionals of different genders and ethnicities with a broad range of technical expertise. Below are short biographies of the organizers:

\begin{itemize}
\itemsep0em
\item \textbf{Amir Rasouli} is a Senior Staff Scientist leading the behavior prediction team at Huawei Noah's Ark Lab. He received his Ph.D. degree in Computer Science from York University in 2020. His current work focuses on various aspects of road user behavior prediction including analysis of naturalistic driving data, understanding theoretical foundations of interactions between heterogeneous traffic participants, modeling vehicle and pedestrian behavior, simulation, and development of predictive models for intelligent driving systems.

\item \textbf{Randy Goebel} is a Professor of Computing Science at the University of Alberta, and a co-founder of the Alberta Machine Intelligence Institute (Amii), one of the three Pan Canadian AI Institutes (with MILA and Vector). Relevant to AI and autonomous driving (AD), he has spent a year with the Volkswagen Group Global AI Lab in Munich, and is a scientific advisor to the German Research Centre for AI (DFKI).  His work on formalizing explainable AI (XAI) is now being integrated into a framework for explainable reinforcement learning (XRL) as a foundation for regulatory compliant AD.

\item \textbf{Matthew E. Taylor} is an Associate Professor of Computing Science at the University of Alberta and a Fellow-in-Residence and Canada CIFAR AI Chair at Amii. He is also an Adjunct Professor at Washington State University and was formerly the Principal Researcher at Borealis AI in Edmonton, the artificial intelligence research lab for the Royal Bank of Canada. His research focuses on developing intelligent agents that interact with their environments to enable individual agents, and teams of agents, to learn tasks in real-world environments that are not fully known when the agents are designed; to perform multiple tasks, rather than just a single task; and to robustly coordinate with, and reason about, other agents.

\item \textbf{Iuliia Kotseruba} is a Ph.D. candidate at York University. She completed her B.Sc. in Computer Science at University of Toronto in 2010 and M.Sc. in Computer Science at York University in 2016.  
Her research interests include human visual attention, computer vision, and human behavior understanding for applications in systems for assistive and autonomous driving.
\item \textbf{Tianpei Yang} is a Postdoctoral Researcher at Intelligent Robot Learning Lab at the University of Alberta. She received her Ph.D. from Tianjin University in 2021. Her research focuses on multi-agent systems and deep reinforcement learning, particularly on facilitating efficient and scalable RL through transfer learning, reinforcement learning, hierarchical RL, opponent modeling, and norm emergence in multi-agent systems.

\item \textbf{Soheil Alizadeh} is a Senior Engineer at Huawei Noah's Ark Lab. He received his Ph.D. degree from University of Toronto in 2019. His research interests are in machine learning, reinforcement learning, control and optimization. His current work focuses on autonomous driving, traffic control, and traffic simulation applications.

\item \textbf{Montgomery Alban} is a Senior Software Engineer at Huawei Noah's Ark Lab leading the development of the open-source simulation environment SMARTS, particularly focusing on the platform design to satisfy the research objectives of users, software development coordination, and system administration. 
\item \textbf{Florian Shkurti} is an Assistant Professor of Computer Science at University of Toronto and a fellow of the NSERC Canadian Robotics Network (NCRN). He earned his Ph.D. degree in Computer Science \& Robotics at McGill University in 2018. His research is at the intersection of autonomous mobile robotics, machine learning, computer vision, and control. He is interested in decision-making and control under uncertainty, imitation learning, inverse reinforcement learning, autonomous visual exploration and search.

\item \textbf{Yuzheng Zhuang} is a Senior Research Scientist from the reinforcement learning research team at Huawei Noah’s Ark lab HQ. She received her B.Sc. in Mathematics from the University of Washington and a M.Sc. in Data Science from New York University. Her research focuses on data-efficient reinforcement learning and imitation learning for autonomous driving planning and control in simulation.

\item \textbf{Adam Scibior} is CTO and co-founder of Inverted AI where he directs its multi-agent behavioral modeling for autonomous driving research.  He also is an Adjunct Professor of Computer Science at UBC and received his Ph.D. from Cambridge University. His research focuses on probabilistic programming, in particular on exploiting composition of Bayesian inference algorithms. 

\item \textbf{Kasra Rezaee} is a Senior Research Engineer at Huawei Noah’s Ark Lab. He received his Ph.D. degree in Civil Engineering focused on intelligent transportation systems from University of Toronto. His work involves research and development of planning and decision making solutions for self-driving cars, with focus on learning-based approaches. His expertise is in advanced control systems, machine learning, reinforcement learning, intelligent transportation systems, and embedded systems.

\item \textbf{Xinyu Wang} is ADS Planning and Control Chief Architect at Huawei IAS BU. He received his Ph.D. and B.Sc. degrees from Beijing Institute of Technology. His research interests include robotics and autonomous vehicles.

\item \textbf{Weinan zhang} is Associate Professor at Shanghai Jiao Tong University. He earned his Ph.D. from University College London in 2016 and B.Eng. from ACM Class of Shanghai Jiao Tong University in 2011. His research interests include (multi-agent) RL, deep learning and data science with various real-world applications of recommender systems, search engines, text mining and generation, knowledge graphs, game AI, etc.

\item \textbf{Xi Chen} is a Senior Principle Researcher at Huawei Noah's Ark Lab, Montreal, leading a team on AI/ML applications for networks. He also serves as an Adjunct Professor at School of Computer Science, McGill University. His experience and passion extend to a wide range of AI domains, including AI for 5G/6G, AI for communications, prediction, smart IoT, WiFi sensing, NLP, self-driving, smart homes, smart systems, vehicle-to-everything, etc. He obtained his Ph.D. degree at School of Computer Science, McGill University.

\end{itemize}

\subsection{Advisers}
A number of leading ML researchers from top Canadian Universities serve as advisers. Their short biographies are listed below: 
\begin{itemize}
\itemsep0em
\item \textbf{Animesh Garg} is a CIFAR Chair Assistant Professor of Computer Science at the University of Toronto and a faculty member at the Vector Institute where he leads the Toronto People, AI, and Robotics (PAIR) research group. He also spends time as a Senior Researcher at Nvidia Research in ML for Robotics. He earned a Ph.D. in Operations Research from UC Berkeley. His current research focus is on developing algorithmic foundations for Generalizable Autonomy in robotics which involves a confluence of representations and algorithms for reinforcement learning, control, and perception.

\item \textbf{David Meger} is an Assistant Professor of Computer Science at McGill University and a fellow of the NSERC Canadian Robotics Network (NCRN). He received his Ph.D. degree in Computer Science from University of British Columbia. His research interests lie in autonomous robotics, visual attention, planning, perception, navigation, and visual search. He is specialized in machine learning techniques including reinforcement learning, imitation learning, adversarial learning methods, and deep learning. 

\item \textbf{Jun Luo} is a Distinguished Researcher at Huawei Noah's Ark Lab. He studied Computer Science at Peking University and has a Ph.D. in Computer Science and Cognitive Science from Indian University Bloomington. He previously served as an Assistant Professor of Cognitive Science at the University of Toronto. He joined Huawei Technologies in 2016. Since then he has been doing research at the intersection of autonomous driving and reinforcement learning, including serving as the main architect of the SMARTS open-source platform.

\item \textbf{Liam Paull} is an Assistant Professor at University of Montreal and the head of the Montreal Robotics and Embodied AI Lab (REAL). He is a co-founder and director of the Duckietown Foundation, which is dedicated to making engaging robotics learning experiences accessible to everyone. He obtained his Ph.D. from the University of New Brunswick in 2013. His research focuses on robotics problems including building representations of the world, such as for SLAM, modeling of uncertainty, and building better workflows to teach robotic agents new.

\end{itemize}

\subsection{Resources provided by organizers, including prizes}

\begin{description}
\itemsep0em
\item \textbf{Resources.} We will provide a website for the participants to download the software and required documentation and to submit their solutions. Dedicated support staff for the competition will assist participants in using the software and tools provided. 

\item \textbf{Sponsor.} Huawei Technologies Canada, Noah's Ark Laboratory

\item \textbf{Prizes.} We offer the following prizes to the winners in each track:
\begin{itemize}
\itemsep0em
\item Gold: US\$6000  + certificate
\item Silver: US\$4000  + certificate 
\item Bronze: US\$2000  + certificate 
\end{itemize}

Additional prizes will be offered to encourage innovation and participation:
\begin{itemize}
\itemsep0em
\item US\$1000 for the most innovative approach out of top-6 finalists in both tracks
\item US\$1000 given to one of the valid contributions among those that were not in  the top-3 positions in either track
\end{itemize}
\end{description}

\bibliographystyle{abbrv}
\bibliography{driving_smarts}

\begin{thebibliography}{1}

\bibitem{kovvali2007video}
V.~G. Kovvali, V.~Alexiadis, and L.~Zhang.
\newblock Video-based vehicle trajectory data collection.
\newblock In {\em Transportation Research Board Annual Meeting}, 2007.

\bibitem{Sun_2020_CVPR}
P.~Sun, H.~Kretzschmar, X.~Dotiwalla, A.~Chouard, V.~Patnaik, P.~Tsui, J.~Guo,
  Y.~Zhou, Y.~Chai, B.~Caine, V.~Vasudevan, W.~Han, J.~Ngiam, H.~Zhao,
  A.~Timofeev, S.~Ettinger, M.~Krivokon, A.~Gao, A.~Joshi, Y.~Zhang, J.~Shlens,
  Z.~Chen, and D.~Anguelov.
\newblock {Scalability in Perception for Autonomous Driving: Waymo Open
  Dataset}.
\newblock In {\em Conference on Computer Vision and Pattern Recognition
  (CVPR)}, 2020.

\bibitem{zhou2020smarts}
M.~Zhou, J.~Luo, J.~Villella, Y.~Yang, D.~Rusu, J.~Miao, W.~Zhang, M.~Alban,
  I.~Fadakar, Z.~Chen, A.~C. Huang, Y.~Wen, K.~Hassanzadeh, D.~Graves, D.~Chen,
  Z.~Zhu, N.~Nguyen, M.~Elsayed, K.~Shao, S.~Ahilan, B.~Zhang, J.~Wu, Z.~Fu,
  K.~Rezaee, P.~Yadmellat, M.~Rohani, N.~P. Nieves, Y.~Ni, S.~Banijamali, A.~C.
  Rivers, Z.~Tian, D.~Palenicek, H.~bou Ammar, H.~Zhang, W.~Liu, J.~Hao, and
  J.~Wang.
\newblock {SMARTS: Scalable Multi-Agent Reinforcement Learning Training School
  for Autonomous Driving}.
\newblock In {\em Conference on Robot Learning (CoRL)}, 2020.

\end{thebibliography}

\end{document}